\documentclass[sigconf]{acmart}

\usepackage{multirow}
\usepackage{makecell}

\AtBeginDocument{%
  \providecommand\BibTeX{{%
    \normalfont B\kern-0.5em{\scshape i\kern-0.25em b}\kern-0.8em\TeX}}}

\setcopyright{acmcopyright}
\copyrightyear{2024}
\acmYear{2024}

\acmConference[HRI '24]{Taking a Closer Look: Refining Trust and Its Impact in HRI Workshop}{March 11--15, 2024}{Boulder, CO}
\begin{document}

\title{Measuring Trust for Exoskeleton Systems}

\author{Leia Stirling}
\email{leias@umich.edu}
\authornotemark[1]
\orcid{0000-0002-0119-1617}
\affiliation{%
  \institution{University of Michigan}
  \city{Ann Arbor}
  \state{MI}
  \country{USA}
  \postcode{48109}
}

\author{Man I Wu}
\email{maniwu@umich.edu}
\affiliation{%
  \institution{University of Michigan}
  \city{Ann Arbor}
  \state{MI}
  \country{USA}
  \postcode{48109}
}

\author{Xiangyu Peng}
\email{xypeng@umich.edu}
\affiliation{%
  \institution{University of Michigan}
  \city{Ann Arbor}
  \state{MI}
  \country{USA}
  \postcode{48109}
}

\renewcommand{\shortauthors}{Stirling et al.}

\begin{abstract}
Wearable robotic systems are a class of robots that have a tight coupling between human and robot movements. Similar to non-wearable robots, it is important to measure the trust a person has that the robot can support achieving the desired goals. While some measures of trust may apply to all potential robotic roles, there are key distinctions between wearable and non-wearable robotic systems. In this paper, we considered the dimensions and sub-dimensions of trust, with example attributes defined for exoskeleton applications. As the research community comes together to discuss measures of trust, it will be important to consider how the selected measures support interpreting trust along different dimensions for the variety of robotic systems that are emerging in the field in a way that leads to actionable outcomes.
\end{abstract}

%


\keywords{wearable robot, trust, direct measures, user perceptions}


\received{2 February 2024}
\received[accepted]{16 February 2024}

\maketitle

\vspace{-2mm}
\section{Introduction}

Robotic systems are being designed to perform many different types of tasks, including those not suited for direct human involvement, where robots operate independently under human supervision (\textit{e.g.,} risky environments), tasks that require the robot to work safely in close proximity and in collaboration with humans (\textit{e.g.,} synchronized tasks, co-manipulation), and tasks where the robot is wearable and must align with the human movement (\textit{e.g.,} exoskeletons, prosthetics). In each of these scenarios, it is important to measure the trust a person has that the robot can support achieving the desired goals. Law and Shultz \cite{law2021trust} surveyed measures of trust for human-robot interactions, highlighting measures of perceived trust (through surveys) and direct measures (through observed actions). However, the development of trust measures has predominantly focused on robotic systems that are not wearable, that is the robot can move separately from the human. As we build out methodologies for assessing trust, we must consider the role of the robot to support defining and selecting relevant metrics of trust.

While some measures of trust may apply to all potential robotic roles, there are key distinctions between wearable and non-wearable robotic systems. Robots will naturally have behaviors that are influencing the final task goal. If a non-wearable robot takes an action that is supportive or contrary to the goal, the human can change their behavior to accommodate the robot's behavior and work towards the goal. However, in a wearable robot, the actions of the human and robot must be coordinated. Rather than adapting to the robot's behavior with complementary actions, the human user must actively correct the robot's behavior. Failure to do so could result in the wearable robot inadvertently misguiding the human's actions, leading to higher effort, increased injury risk, and/or inability accomplish the task. While the consideration of motion planning and robot decisions is always important to achieve the goal for a human-robot team, the human modulation of their behavior can be more constrained with the wearable system. If we specifically consider the human as a supervisor of a non-wearable robot without co-manipulation, it is possible that humans will miss the robot's errors. If the robotic system is not worn, but has co-manipulation of an object, errors can be perceived through interaction forces leading to more synchronized behaviors \cite{nikolaidis2017human}. However, wearable robots such as exoskeletons have a tight coupling with the human movement, directly applying forces and torques to the user. The threshold to perceive an exoskeleton error in timing for an ankle exoskeleton has been shown to be as low as 2.8\% of a stride period \cite{Peng2022}. Thus, exoskeleton errors may be more easily perceived than robots that are supervised or are co-manipulating objects, which impacts the effect of motion plan variability on trust.

When a robot fails to meet the user's expectations, trust is impacted. These expectations may include the spatial motion of the robot, the verbal or non-verbal feedback, or the order of a task procedure. It could be that the system is operating as designed, but the user does not understand the rationale. This disconnect between human expectation and system feedback was observed when users operated an active elbow exoskeleton controlled by electromyography (EMG) signals, with EMG feedback presented to the user \cite{peng2023effects}. Some participants believed they had fully relaxed their muscles when their muscles still maintained a low activation. This illusion of complete relaxation was observed in situations like operating against gravity, where slight muscle contraction is normally needed, but was not necessary with the exoskeleton's presence. This misalignment between the programmed exoskeleton controller and the natural human motor control could lead to a decrease in trust when using the exoskeleton.

Another example of not meeting expectations is when a robot fails to perform a known action that should be taken. When a lower extremity exoskeleton has an error, such as a missed actuation during walking, there are immediate effects on the biomechanics during the step of the missed actuation \cite{WuMA1}. The frequency of errors can then influence and lead to compensatory mechanisms even when the exoskeleton is behaving nominally \cite{WuMA2}. We have examined dimensions of trust for exoskeletons through perceived predictability and supportiveness \cite{Wu2023IROS}. Validity was examined using fixed error rates of the ankle-torque controller and these perceived measures. As hypothesized, there were correlations supporting a decrease in perceived predictability and supportiveness with increasing error rates \cite{Wu2023IROS}. 

For wearable systems, the dimensions of trust should be specified to consider relevant influences on the construct of trust and the surrogate measures used to quantify trust. In this paper, we consider current measures of trust and extensions to support measuring trust in tightly-coupled wearable robots. Section \ref{sec:current trust} presents a selection of current trust measures, Section \ref{sec:exo attributes} creates specific trust attributes for exoskeleton applications, and Section \ref{sec:validity} introduces considerations on validity for trust attributes. Through this paper, we bring the consideration of wearable robotic systems to the conversation of trust measurements in robotics.

\vspace{-2mm}
\section{Current Trust Attributes}
\label{sec:current trust}
An accepted definition for trust in human-automation-robotic systems is ``the attitude that another entity (e.g., human, machine, system) will help achieve a person’s goals in a situation characterized by uncertainty and vulnerability'' \cite{Lee2004}. We can operationalize this construct by defining its dimensions and sub-dimensions, which are ways we can observe or learn about the concept from a specific interpretation or facet of the construct. We can then define attributes, which are measurements we can make to quantify the dimension. Lee and See \cite{Lee2004} define three dimensions of trust -- Purpose, Process, and Performance. In the context of trust, purpose is the understanding of why the system was developed, process is an understanding of how the system operates, and performance is an understanding of what the system can do. We can then define measurable attributes (both human perception and direct measures) aligned with these dimensions.  

Jian et al.~\cite{jian2000foundations} approached defining attributes of trust for general automated systems through ethnographic studies to identify words related to trust and distrust. Their attributes can be considered in context with the dimensions of Lee and See \cite{Lee2004} and include ratings of perceived deception in the system, wariness to use the system, if the system leads to injurious outcomes, confidence in system, if the system provides security, if the system has integrity, if the system is dependable, if the system is reliable, and a general trust in the system. A challenge with this framework when applied to a robotic system is that robotic systems have multiple components, each which may influence trust differently. Yagoda and Gillan \cite{yagoda2012you} developed a trust scale that includes similar attributes of reliability, dependability, understandability, and accessibility, but applies them individually to sub-systems of the HRI system (\textit{e.g.}, user interface, sensor data, navigation capabilities, signal bandwidth, end effectors, remote information processing, level of automation, type of control). This structure provides additional context for interpreting perceptions of trust in robotic systems. 

However, user perceptions are not the only way that trust is observed. Law and Scheutz \cite{law2021trust} describe different direct measures of trust falling into categories of task intervention (\textit{e.g.}, does the human takeover), task delegation (\textit{e.g.}, what tasks are assigned to the robot), behavioral change (\textit{e.g.}, modifications of behavior related to supporting safety, changing actions that were taken without the system), and following advice (\textit{e.g.}, is the human following the robot suggestion). Through these observed behaviors, we can also learn how the person actually used a system.

With user perceptions and direct measures, researchers have investigated how adapting the environment (inclusive of tasks) and human characteristics influence trust by manipulating the robotic system, adapting the environment or task, and considering different participant populations \cite{hancock2011meta,law2021trust}. By modulating these additional factors, studies can determine how to generate requirements, define operational design domains, redesign robotic systems, and develop training methods. While these factors are not dimensions of trust, they are important factors that influence trust. 

Wearable robotic devices are designed to support goals and tasks; however, wearable robots also physically interact with the human such that movements of the human need to align with the robot and the robot needs to align with the human. This tight coupling introduces unique attributes in comparison to non-wearable robotic systems that align with the overarching dimension of purpose, process, and performance.

\begin{table*}[]
\centering
\caption{Attributes for assessing trust dimensions from user perceptions}
\label{tab:userperception}
\begin{tabular}{p{18mm}|p{25mm}|p{60mm}|p{60mm}}
Dimension & Sub-Dimension & Description  & Example Attribute  \\
\hline
\multirow{2}{*}{Purpose} & Motives/Intentions & The degree to which the intentions of the system are aligned with the goals of the user &  \textit{Rate the scenarios for which you would use the exoskeleton.}\\ \cline{2-4}
                         & Faith/Wariness & The degree to which a user believes that a system can achieve a goal   &  \textit{Rate how your feel the exoskeleton enables <task>. }\\
\hline
\multirow{2}{*}{Process} & Accessibility & The degree to which information is available. & \textit{The task feedback is accessible when needed.} \\  \cline{2-4}
                         & Integrity & The degree to which the system adheres to principles of the user.    & \textit{Rate how you feel the exoskeleton supports or limits motor actions.}\\ \cline{2-4}
                         & Dependability & The degree to which the system's behavior is consistent.    & \textit{Rate your trust in the exoskeleton activating at the right time during the task.}\\ \cline{2-4}
                         & Understandability & The degree to which the user is able to interpret the system's operation.  & \textit{Rate whether the exoskeleton actions align with your expectations.}\\ \cline{2-4}
                         & Familiarity & The degree to which the user is able to manipulate the system.  & \textit{Rate how often you use this exoskeleton.}\\
\hline
\multirow{2}{*}{Performance} & Ability & The degree to which the system achieves a specific goal. &\textit{Rate the responsiveness of the exoskeleton. \;\;\;\; Rate how the exoskeleton helped or hindered you to perform the task.}\\ \cline{2-4}
                             & Reliability (System Confidence) & The degree to which the system consistently meets goals. & \textit{Rate how effectively the exoskeleton determined your intent to move.}  \\   \cline{2-4}
                             & Safety & The degree to which the user feels safe using the system & \textit{Rate how safe you felt while walking with the exoskeleton.}  \\
\hline
\end{tabular}
\end{table*}
\vspace*{-4mm}

\section{Attributes of Trust for Exoskeletons}
\label{sec:exo attributes}
When specifying the attributes to be considered for measuring trust, we should consider what happens when a person's trust is not calibrated with the system to support selecting surrogate measures for trust. If the user perceives the system's capabilities are higher than the true capabilities, the user may over-trust the system and misuse the exoskeleton; for example, using the exoskeleton in tasks or environments for which it is not appropriate. Alternatively, if system capabilities are perceived as lower than they actually are, the human will distrust the exoskeleton, which may lead to inappropriate compensatory mechanisms (\textit{i.e.}, alternate motor strategies) or disuse in situations where the exoskeleton could provide benefit. It is appropriate for a user to have dynamic trust, where in some scenarios they have lower trust and other scenarios they have a higher trust. We want to be able to understand the user's trust across tasks, environments, and timelines, which requires relevant attributes to be defined for user perceptions and direct measures that are surrogates for the dimensions and sub-dimensions of interest.

Key to wearable robots are the ways we frame perceptual questions and directly measured  attributes. In Table \ref{tab:userperception} and Table \ref{tab:directmeasures}, we use the dimensions and sub-dimensions from the literature to generate attributes with additional specificity for an exoskeleton use case. For example, rather than asking about dependability generically, we can be more specific and ask users to rate their trust in the exoskeleton activating at the right time during the task. By providing additional specificity on the attribute of dependability, we can more directly inform design changes in the system if needed. 

There is a flow between performance, process, and purpose. Within performance are the characteristics of the task as they occur. As noted by Lee and See \cite{Lee2004}, these experiences lead to expectations on process, which are ``qualities and characteristics attributed to the agent.'' From the expectations, a person develops their faith in the system and intentions to use the system. This differentiation informs the question framing (Table \ref{tab:userperception}). We can create an ability attribute that considers a specific task by asking users to rate the responsiveness of the exoskeleton, or to rate how the exoskeleton helped or hindered a specific task. This ability grows to an expectation on system integrity, where the question is framed as rating how the exoskeleton generally supports or limits motor actions. This expectation then leads to intention, where we ask users to rate the scenarios they would use the exoskeleton. Providing additional specificity in user perceptual questions beyond the sub-dimension label can limit the assumptions users may make when responding to these questions and can enable more actionable outcomes.

We can also define surrogate direct measures aligned with the tight coupling of the robot to the human motion. For example, behavioral changes include monitoring compensatory mechanisms, such as increased peak hip flexion during gait (lower extremity exoskeleton) or increased peak shoulder flexion (upper extremity exoskeleton). For exoskeleton systems, task intervention can be interpreted as trying to oppose the system's selected action, which can be directly measured through antagonistic muscle activation or interaction forces between the user and exoskeleton. To infer changes in directed attention or understandability using direct measures, gaze tracking features can be estimated (\textit{e.g.}, scan patterns, dwell times, visit frequencies). Confusion may be observed through increased visit frequency to a display or to the system itself, or increased dwell times to these locations. For direct measures of purpose, rather than asking questions of intention to use, actual usage in deployed settings can be monitored to determine what tasks are attempted with the exoskeleton and in which environments. For example, perhaps a person uses an exoskeleton to walk (the task) on sidewalks (the environment), but does not use the exoskeleton to walk on hiking trails. Current exoskeleton test deployments in industry highlight that even when the exoskeleton could be used, workers are not always using the system. 

By combining measures of purpose, process, and performance through user perceptions and direct measures, we can gain a deeper understanding of the sub-dimensions that are influencing overall system trust and usage. Once we have sub-dimensions defined, providing users additional context on user perception questions and creating surrogate direct measures can provide a richer understanding of system trust.

\begin{table*}[]
\centering
\caption{Attributes for assessing trust dimensions from direct measures}
\label{tab:directmeasures}
\begin{tabular}{p{18mm}|p{25mm}|p{60mm}|p{60mm}}
Dimension & Sub-Dimension & Description & Example Attribute   \\
\hline
\multirow{2}{*}{Purpose} & Task delegation  & Does the person use the robot for its intended tasks or tasks outside of its scope?   & \textit{Tasks attempted with exoskeleton support}\\ \cline{2-4}
                         & Usage environment & Is the robot used in its intended environment?   & \textit{Environments where exoskeleton is used} \\
\hline
\multirow{2}{*}{Process} & Understandability & The degree to which the user is able to interpret the systems' operation.  & \textit{Gaze tracking features for confusion monitoring} \\ 
\hline
\multirow{2}{*}{Performance} & Task intervention & Does the person override the robot's actions to intervene with the task? & \textit{Antagonistic muscle activation, interaction forces} \\   \cline{2-4}
                             & Behavioral change & Does the user modify their actions in the presence of the system? & \textit{Peak hip flexion; Peak shoulder flexion}    \\ \cline{2-4}
                             & Advice following  & Does the user follow the robot's actions in context of the task and environment?  &\textit{Muscle co-contractions}  \\ \cline{2-4}
                             & Directed attention   & Does the person change their focus when interacting with the robot? & \textit{Gaze tracking features for areas of interest}\\
\hline
\end{tabular}
\end{table*}
\vspace*{-5mm}

\section{Considering Validity}
\label{sec:validity}
We can think about trust metric validation from a few different perspectives, including content validity and performance characteristics (also termed criteria validity) \cite{hulley2007designing,strauss2009construct}.

Content validity considers whether all aspects of the construct are accounted for with the metric. While content validity has long been debated,  ``it is concluded that although measures cannot be `validated' based on content validity evidence alone, demonstration of content validity is a fundamental requirement of all assessment instruments'' \cite{sireci1998construct}. As seen in Tables \ref{tab:userperception} and \ref{tab:directmeasures}, trust is composed of multiple dimensions. There is no single value that will describe all elements of trust. Through the use of multiple attributes, we can make inferences for the questions of interest. While one option is to create a composite metric combining the underlying attributes into an overall trust score, we believe it is useful to separately consider attributes and their associated dimensions as the individual metrics could lead to different design choices. For example, lower trust on process attributes may require different solutions than lower trust on performance attributes. 

Construct validity consists of both convergent and discriminant validity. As the attributes defined are abstractions that map to trust dimensions, it is important that we understand their benefits and limitations. Convergent validity can be assessed by collecting more than one measure for a particular dimension to assess similarity. Having redundancy in measures through perceptions and direct measures also builds confidence that there is evidence a sub-dimension is important for a specific system. Discriminant validity suggests that the measure is unrelated or negatively related to measures of distinct constructs. However, for exoskeletons, we can see that compensatory movements can be surrogates for trust, but are also aligned with constructs of movement coordination. An increase in trust should align with an increase in movement coordination. Yet, movement coordination can increase or decrease separately from trust in the exoskeleton. We would not use movement coordination to discriminate between trusters and non-trusters alone, but it can be used within a repeated measures framework to examine how a user operates with the system. Rather than strictly require discriminant validity, researchers should consider additional factors that may influence the attribute for inferences made.   

\vspace{-4mm}
\section{Discussion}
\label{sec:discussion}
In this paper, we have considered the dimensions and sub-dimensions of trust, with example attributes defined for exoskeleton applications. As trust can be decomposed into several relevant dimensions, it may not be enough to support the design iteration process to have a single overall trust composite metric or a general rating on a sub-dimension term directly. By developing attributes that have additional specificity, we can improve design iterations, but also limit the variability in assumptions users make when responding to these perception-based questions. Collecting measures of both user perceptions and actual behaviors will be important to support the validity of the measures selected and the inferences made. 

There are many different methods for measuring trust in the literature and as researchers, we may be inclined to pick and choose the attributes that seem relevant for our study. However, Chita-Tegmark et al.~\cite{chita2021can} caution that removing questions can lead to missing information about individual differences in the characteristics we attribute to robots. So there is a balance between selecting the relevant attributes for a specific study with the ability to detect individual differences in users for a specific robot-task-environment combination. As the research community comes together to discuss measures of trust, it will be important for those measuring trust to consider how the selected measures  support interpreting trust along different dimensions for the variety of robotic systems that are emerging in the field in a way that leads to actionable outcomes.
\vspace{-4mm}

\section{Acknowledgements}
This work was supported by the National Science Foundation under Grant 1952279 (MW, LS) and Grant 2110133 (XP, LS).

\vspace{-2mm}

\bibliographystyle{plain}
\bibliography{references.bib}

\end{document}